\documentclass{article}

\usepackage{arxiv}

\usepackage[utf8]{inputenc} 
\usepackage[T1]{fontenc}    
\usepackage{hyperref}       
\usepackage{url}            
\usepackage{booktabs}       
\usepackage{amsfonts}       
\usepackage{nicefrac}       
\usepackage{microtype}      
\usepackage{lipsum}		
\usepackage{graphicx}
\usepackage{natbib}
\usepackage{doi}

\usepackage{cite}
\graphicspath{{figures/}} 

\usepackage{boites}
\usepackage{amsmath}
\usepackage{amssymb,bm}
\usepackage{cite}
\usepackage{mathtools}
\usepackage{amsfonts}       
\usepackage{soul,kpfonts}
\usepackage[ruled,linesnumbered]{algorithm2e}
\usepackage{subfig,natbib}
\usepackage{multirow}
\usepackage{dcolumn}
\title{Gravity aided navigation using Viterbi map matching algorithm}

\author{ {\hspace{1mm}Wenchao Li}\\
	School of Science\\
	RMIT University\\
	Melbourne, VIC, Australia \\
	\texttt{wenchao.li@rmit.edu.au} \\
	\And
	{\hspace{1mm}Christopher Gilliam}\\
	School of Engineering\\
	RMIT University\\
	Melbourne, VIC, Australia \\
	 \\
    \And
	{\hspace{1mm}Xuezhi Wang}\\
	School of Science\\
	RMIT University\\
	Melbourne, VIC, Australia \\
	 \\	
    \And
	{\hspace{1mm}Allison Kealy}\\
	School of Science\\
	RMIT University\\
	Melbourne, VIC, Australia \\
	 \\		 
    \And
	{\hspace{1mm}Andrew D. Greentree}\\
	Australian Research Council Centre of\\ Excellence for Nanoscale BioPhotonics\\
	School of Science\\
	RMIT University\\
	Melbourne, VIC, Australia \\
	 \\		 
    \And
	{\hspace{1mm}Bill Moran}\\
	Department of Electrical and\\ Electronic Engineering\\
	University of Melbourne\\
	Melbourne, VIC, Australia \\
}

\date{}

\renewcommand{\shorttitle}{Navigation using Viterbi Map Matching}

\begin{document}
\maketitle

\begin{abstract}
	In GNSS-denied environments, aiding a vehicle's inertial navigation system (INS) is crucial to reducing the accumulated navigation drift caused by sensor errors (e.g. bias and noise). One potential solution is to use measurements of gravity as an aiding source. The measurements are matched to a geo-referenced map of Earth's gravity in order to estimate the vehicle's position. In this paper, we propose a novel formulation of the map matching problem using a hidden Markov model (HMM). Specifically, we treat the spatial cells of the map as the hidden states of the HMM and present a Viterbi style algorithm to estimate the most likely sequence of states, i.e. most likely sequence of vehicle positions, that results in the sequence of observed gravity measurements. Using a realistic gravity map, we demonstrate the accuracy of our Viterbi map matching algorithm in a navigation scenario and illustrate its robustness compared to existing methods.

\end{abstract}


\section{Introduction}
Since the advent of Global Navigation Satellite Systems (GNSS) in the 1960s, GNSS-based positioning has become ubiquitous in the areas of navigation, guidance and control. However, in a growing number of situations GNSS is either unavailable, e.g. in underwater environments~\citep{wang2017location}, or deliberately jammed or spoofed~\citep{hemann2016long}. In such GNSS-denied (or contested) situations, a platform's navigation performance is governed by on-board sensors -- chief among them being the inertial navigation system (INS), which integrates inertial measurements from accelerometers and gyroscopes to obtain position. Unfortunately, even with high precision INSs, very small errors in the inertial sensors result in the build up of large navigation errors over time~\citep{Titterton2004,Groves2013}. Thus, an on-board aiding source that provides a position fix is crucial to removing these accumulated navigation errors and retaining confidence in the navigation accuracy. A promising candidate, and the focus of this paper, is the use of gravitational information to obtain a position fix~\citep{han2016improved,liu2019navigability,xuezhi2022}.

Gravity aided navigation originates from the early 1990's~\citep{Affleck1990,Jircitano1991}. The central concept is that an on-board sensor measures elements of the gravitational vector (or gravity gradient tensor) whilst the platform is in motion and these measurements are then matched to a geo-referenced map of the Earth's gravitational field to determine a position. The advantage of such an approach is twofold: First, gravity is an inherent physical property of the Earth, thus it is immune to environmental interference, jamming and spoofing. Second, the measurement of gravity is a passive operation that does not require external information, which is advantageous in sensitive situations~\citep{muniraj2017framework}. The challenge in this approach however is how to match the measurements to a position in the map. The design of this matching procedure, known as the map matching algorithm, is key to the success of gravity aided navigation.

The map matching algorithm needs to take into account: 1) the sensor noise when measuring gravity; 2) the uncertainty on the spatial location when a measurement is obtained; 3) possible ambiguity in the map resulting in a measurement being matched to multiple locations. In the literature, map matching algorithms tackle these problems in one of two ways. The first category of algorithms consider the problem as a single point matching operation. Building on the framework introduced in the Sandia terrain-aided navigation system (SITAN)~\citep{hostetler1983nonlinear,bergman1997bayesian}, these techniques use the geo-referenced map as a look-up function and compute a predicted gravitational measurement using the INS estimated position. The predicted measurement is then used in the prediction step of a statistical filter. Due to the non-linear relationship between the estimation states and gravitational measurements, the extended Kalman filters (EKF) is often used to perform the estimation~\citep{wei2017robust,Lee2015}. This category of algorithms take into account the sensor noise and spatial uncertainty but do not consider the map ambiguity directly. The second category of map matching algorithms directly considers the structure of the map. They approach the problem as a sequential matching operation. The classical example is the iterative closest contour point (ICCP) algorithm proposed in~\citep{kamgar1999vehicle} and utilised in~\citep{Wang2016IEEE,han2017combined,Liu2019}. The ICCP is based on the observation that a single, scalar, gravity measurement creates an iso-contour of similar values in the gravity map space. Thus, given a sequence of gravity measurements, the matching problem is posed in terms of fitting a trajectory to a set of iso-contours based on initial position estimates and sensor measurements. To make the problem well posed, a regularization term is introduced to control the shape of the trajectory. However, linking the kinematic constraints of the platform’s motion to the regularization term is not straightforward.

In this paper, we propose a novel sequential map matching algorithm based on formulating the matching operation as a hidden Markov model (HMM). Specifically, we consider the spatial cells of the map to be the set of all possible hidden states in the model and the platform's trajectory to be a particular sequence of these states. The observations from each state are the gravitational measurements and the output and state transition probabilities relate to the sensor noise and platform movement models, respectively. Thus, the map matching problem is equivalent to determining this sequence of states given a sequence of gravitational measurements. The advantage to this formulation is twofold: First, our proposed approach works directly in the map space and incorporates the modelling of the sensor noise and platform motion with the underlying structure of the gravity map. Second, this type of HMM problem can be solved efficiently using a dynamic programming method known as the Viterbi algorithm~\citep{Viterbi1967,ren2009hidden}. Given the HMM formulation, the Viterbi algorithm estimates the most likely sequence of states that results in the sequence of observations and this sequence of states is optimal in the maximum a posterior (MAP) sense. Accordingly, we present a Viterbi map matching algorithm that outputs the most likely trajectory of the platform given a sequence of gravitational measurements. Furthermore, to tackle limited map resolution, we implement a two-layer, coarse-to-fine, estimation scheme to achieve sub-cell accuracy. We validate our algorithm hsing the ultra-high resolution, non-parametric, gravity maps presented in~\citep{hirt2013new}  and demonstrate that it outperforms existing algorithms in terms of robustness and navigational accuracy. In particular, we demonstrate that our algorithm is able to tackle varying map spatial resolution as well as varying sensor noise levels.

\section{Problem Formulation}\label{sec_formulation}
Suppose that the actual position of a vehicle at time $k$ is $\mathbf{s}_{k}=[x_k,y_k]^T$, the INS position estimate is $\mathbf{s}_{k}^{INS}=[x_k^{INS},y_k^{INS}]^T$ and the position corrected via map matching algorithm is $\mathbf{s}_{k}^{C\text{-}INS}=[x_k^{C\text{-}INS},y_k^{C\text{-}INS}]^T$. The gravity measurement at time $k$ and position $\mathbf{s}_{k}$ is modeled as
\begin{align}
    z_{k} = g(\mathbf{s}_{k})+\varepsilon_k
\end{align}
where $g(\mathbf{s}_{k})$ describes the true Earth's gravity at position $\mathbf{s}_{k}$ and $\varepsilon_k\sim\mathcal{N}(0,\sigma_z^2)$ is noise with $0$ mean and standard deviation $\sigma_z$.  Without loss of generality, in the simulation, the true Earth gravity is obtained from the gravity map without noise for the purpose of demonstrating the algorithm. Additionally, we assume that the INS  provides velocity measurements $\mathbf{v}_{k}$ in the navigation frame and it is modeled as:
\begin{align}
\mathbf{v}_{k}=\mathbf{v}^0_{k}+\bm{\phi}_{k} \label{vel_mea}
\end{align}
where $\mathbf{v}_{k}^0$ is the ground truth, $\bm{\phi}_{k}$ is the noise term comprising with a bias $\mathbf{b}_k$ and an independent zero mean Gaussian noise with co-variance $\sigma_v^2\mathbf{I}_2$ and $\mathbf{I}_2$ is the $2\times2$ identity matrix. 

We assume that the INS position estimates are corrected after every segment of $T$ gravity measurements. Without loss of generality, we assume that the INS position estimates and measurements are recorded from $k+1$. Then after $T$ sampling times, i.e., at time $k+T$,  we have following available segments of measurements and INS position estimates:
\begin{align}
\mathbf{Z}_k=\{z_{k+1},\cdots,z_{k+T}\},\;
\mathbf{V}_k=\{\mathbf{v}_{k+1},\cdots,\mathbf{v}_{k+T}\},\;
\mathbf{S}^{INS}_k=\{\mathbf{s}_{k+1}^{INS},\cdots,\mathbf{s}_{k+T}^{INS}\}\notag
\end{align}
The corrected position sequence using $(\mathbf{Z}_k,\mathbf{S}^{INS}_k,\mathbf{V}_k)$ and the proposed map matching algorithm is denoted by $\mathbf{S}_{k}^{C\text{-}INS}=\{\mathbf{s}_{k+1}^{C\text{-}INS},\cdots,\mathbf{s}_{k+T}^{C\text{-}INS}\}$. 

In this work, since the map matching algorithm is implemented every $T$ time steps, therefore we actually are interested in how the algorithm works within time slot $\{k+1,k+2,\cdots,k+T\}$. For simplicity, where there is no confusion, time stamp $k$ in this time slot is dropped from now on and we write $t=1,\cdots,T$. 
\section{Viterbi Map Matching Algorithm}\label{algl_sec}

\subsection{Constructing a grid on the gravity map}

The gravity map can be viewed as a grid of pixels with each pixel representing an area equivalent to the resolution of the  map. These pixels will henceforth be referred to as \emph{cells}. The resolution of the gravity map, i.e. the dimensions of the cell, is $R_{x}\times R_{y}$. Since each cell on the gravity map can be associated with a location, we simply let the central position of each cell be its location. Denote the collection of all cells' locations on the interested area of the map by $\bm{\mathcal{C}}$.  We define $g_m(\cdot)$ as the ``look-up function'' for finding the value from the gravity map using the given location, e.g. $g_m(\mathbf{s}_{t}^{INS})$ is the gravity value associated to the cell that $\mathbf{s}_{t}^{INS}$ lies in. 

\begin{figure}[htp!]
    \centering
    \includegraphics[scale=.6]{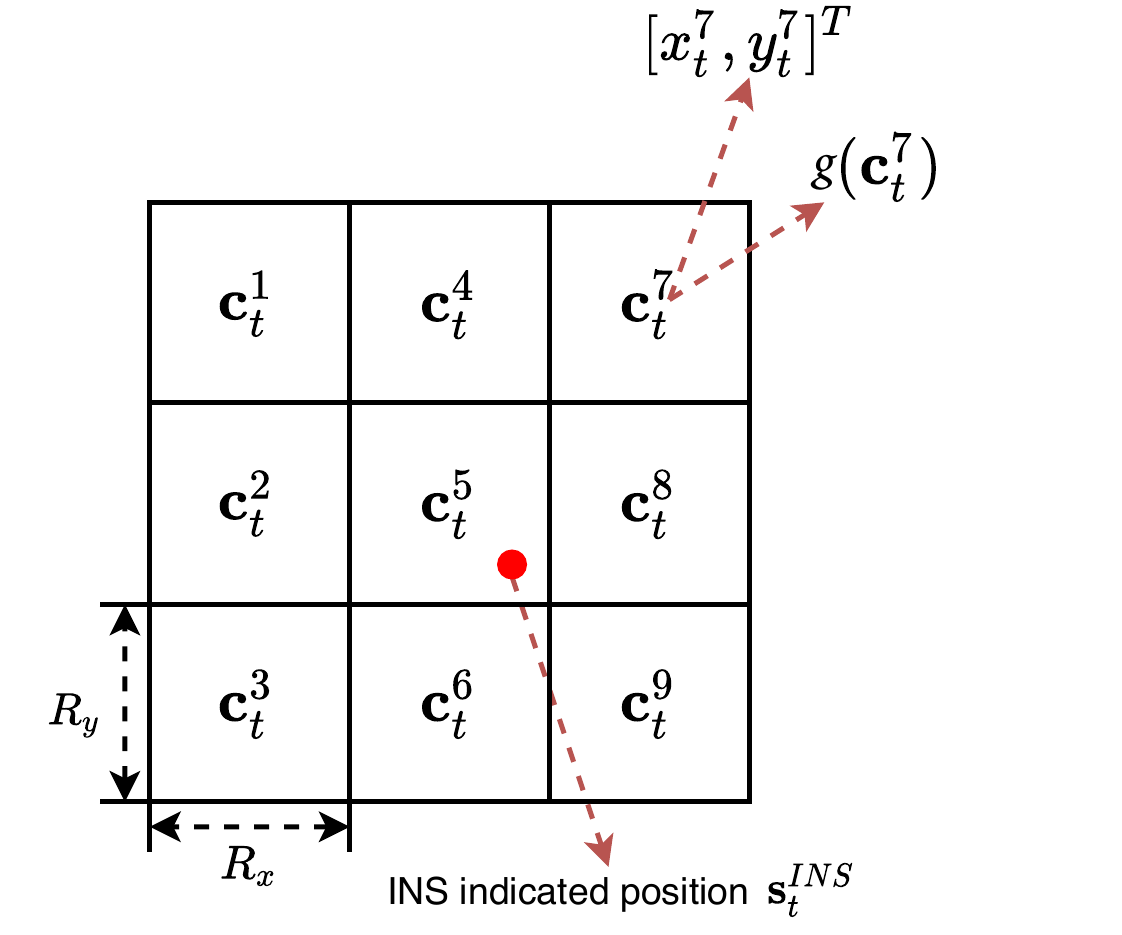}
    \caption{Illustrative example of a $3\times3$ grid of cells, at arbitrary time $t$, from a gravity map $g_m$. Each cell has a spatial location $\mathbf{c}_t^{j}=[x_t^j,x_t^j]^T$ and an associated gravity value $g_m(\mathbf{c}_t^{j})$ at this location. The central cell of the grid, $\mathbf{c}^5_t$, is the one that the INS position estimate $\mathbf{s}_t^{INS}$ lies in and can be determined via~\eqref{eq1}, and the positions of other cells can then be calculated by~\eqref{eq2}. }
    \label{deml_grid}
\end{figure}

In fact, at a specific sampling time $t$,  we are only interested in a limited area around the INS position estimate $\mathbf{s}_{t}^{INS}$ since the true position should be within that region. Therefore, given an estimate $\mathbf{s}_{t}^{INS}$, we can find a cell, or equivalently a pixel, on the geo-referenced map of gravity such that $\mathbf{s}_{t}^{INS}$ lies in. It should be noted that, since the cell is predefined and constructed according to the  pixel of the map, and the estimate $\mathbf{s}_{t}^{INS}$ is arbitrary, therefore $\mathbf{s}_{t}^{INS}$ can be any point within the underlying cell and is unnecessary to be the center position of this cell. Then a grid of cells around the cell containing $\mathbf{s}_{t}^{INS}$ can be constructed and denoted by $\mathbf{B}_t$. Let the size of the grid be $nR_{x}\times nR_{y}$, where $n$ is a relatively small odd integer, and the positions of the cells within this grid be denoted by $\mathbf{c}^{j}_t$ for $j=1,\cdots,n^2$. We also use  $\mathbf{c}^{j}_t$  as the labels of the cells. The cells within the grid are  arranged as shown in Fig.~\ref{deml_grid}, where the  central cell of the grid, labelled as $\mathbf{c}^{\bar{n}}_t$ with $\bar{n}=\frac{n^2+1}{2}$, is the one that $\mathbf{s}_{t}^{INS}$ lies in.  Its position is
\begin{align}
\mathbf{c}^{\bar{n}}_t\triangleq[x^{\bar{n}}_t,y^{\bar{n}}_t]^T
= \arg\min_{[x,y]^T\in\bm{\mathcal{C}}} \left\|[x,y]^T-\mathbf{s}_{t}^{INS}\right\|_2\label{eq1}
\end{align}
Essentially, \eqref{eq1} means that  $\mathbf{c}^{\bar{n}}_t$ is the cell that the INS position estimate $\mathbf{s}_{t}^{INS}$ lies in at time $t$. As a result, each cell is associated to a gravity value according to the gravity map. 
Based on $\mathbf{c}^{\bar{n}}_t$ and $n$, the positions of other cells in the grid $\mathbf{B}_k$ are
\begin{align}
\mathbf{c}^{j}_t&=\mathbf{c}^{\bar{n}}_t-[m_1R_x,m_2R_y]^T,\;\;\text{where}\;\;m_1,m_2=-n,\cdots,n\;\;\text{and}\;\;j=1,\cdots,n^2\label{eq2}
\end{align}

\subsection{Viterbi based map-matching algorithm}\label{subsec_algo}
At each time $t=1,2,\cdots,T$, we construct the grid $\mathbf{B}_{t}$ and its associated cells $\mathbf{c}_{t}^{j_t}$, where the subscript $t$ is again the time index and superscript $j_t$ is the cell label $j_t=1,\cdots,n^2$. Then the problem becomes how to select a cell in each $\mathbf{B}_t$  such that this segment of cells is the optimal estimate of the actual trajectory. 
For a segment of cells,  $\{\mathbf{c}^{j_1}_{1},\cdots,\mathbf{c}^{j_T}_{T}\}$, with corresponding  referenced gravity values, $\{g_m(\mathbf{c}^{j_1}_{1}),\cdots,g_m(\mathbf{c}^{j_T}_{T})\}$,  gravity measurements, $\mathbf{Z}_k$,  INS  information, $\mathbf{V}_k$ and $\mathbf{S}^{INS}_k$, the joint posterior probability is
\begin{align}
p\left(\{\mathbf{c}^{j_1}_{1},\cdots,\mathbf{c}^{j_T}_{T}\}\Big|\mathbf{Z}_k,\mathbf{S}^{INS}_k,\mathbf{V}_k\right)= p\left(\{g_m(\mathbf{c}^{j_1}_{1}),\cdots,g_m(\mathbf{c}^{j_T}_{T})\}\Big|\mathbf{Z}_k,\mathbf{S}^{INS}_k,\mathbf{V}_k\right)
\end{align}
The most probable trajectory from $t=1$ to $t=T$ is given by the MAP estimator
\begin{align}
\{j_1^*,\cdots,j_T^*\}=\arg\max_{{\bm{\mathcal{I}}^{\bigtimes}}}
p\left(\{\mathbf{c}^{j_1}_{1},\cdots,\mathbf{c}_T^{j_T}\}\Big|\mathbf{Z}_k,\mathbf{S}^{INS}_k,\mathbf{V}_k\right)\label{opt1}
\end{align}
where $\bm{\mathcal{I}}=\{1,\cdots,n^2\}$, $\bm{\mathcal{I}}^{\bigtimes}=\varprod_{i=1}^T \bm{\mathcal{I}}$, notation $\varprod$ is Cartesian product and $\{\mathbf{c}^{j_1^*}_{1},\cdots,\mathbf{c}^{j_T^*}_{T}\}$ is the optimal path (locations) of the vehicle estimated from the given measurements. It should be noted that $\mathbf{c}^{j_t}_{t}$ is actually the central position of the each cells of the map; that is, the estimated path for time interval ${1,\cdots,T}$ passes the center of each cell. Finally, the estimated position sequence is then used to reset the corresponding segment of this period estimated from INS. 

The exhaustive searching process is required to find an optimal solution to~\eqref{opt1}, where the computation load of the order of $n^T$ can explode quickly if $n$ and/or $T$ getting large. In this work, we intend to find an efficient algorithm to solve this problem. 

Note that, for arbitrary sequence, $\{j_1,\cdots,j_T\}$
\begin{align}
p\left(\mathbf{c}_{t}^{j_t}\Big|\mathbf{c}_{t-1}^{j_{t-1}},\cdots,\mathbf{c}_{1}^{j_1}\right)=p\left(\mathbf{c}_{t}^{j_t}\Big|\mathbf{c}_{t-1}^{j_{t-1}}\right),\;\;\;\forall t\in\{2,\cdots,T\}\;\;\text{and}\;\;j_t\in\bm{\mathcal{I}},
\end{align}
which implies that the selection of $\mathbf{c}_{t}^{j_t}$ is only dependent on that of $\mathbf{c}_{t-1}^{j_{t-1}}$. This is a typical Markov process~\citep{Stroock2013}. The sequence $\{\mathbf{c}_{1}^{j_1},\cdots,\mathbf{c}_{T}^{j_T}\}$ with the measurements can be modelled by an HMM. The  Viterbi algorithm~\citep{Viterbi1967} is available to compute~\eqref{opt1} and provides a most likely sequence of positions.  

At each time $t$, the likelihood function of $z_t$ given $\mathbf{c}_{t}^{j_t}$ and $g_m(\mathbf{c}_{t}^{j_t})$ is denoted by
\begin{align}
p_z\left(z_t\Big|\mathbf{c}_{t}^{j_t}\right)=p_z\left(z_t\Big|g_m(\mathbf{c}_{t}^{j_t})\right)\label{eq:likelihood_z}
\end{align}
where $p_z(\cdot)$ is the probability density function of the gravity measurement. For each successive pair $\{t-1,t\}$ with $t=2,\cdots,T$, the measurement likelihood function of $\mathbf{c}^{j_{t}}_{t}$ given $\mathbf{v}_{t-1}$, the velocity measurement at time ${t-1}$, and $\mathbf{c}_{t}^{j_{t}}$, is
\begin{align}
p_v\left(\mathbf{c}^{j_{t}}_{t}\Big|\mathbf{v}_{t-1},\mathbf{c}_{t-1}^{j_{t-1}}\right)
\end{align}
where $p_v(\cdot)$ is the probability density function of the velocity measurement.

Then, given $\mathbf{c}^{j_{t-1}}_{t-1}$ and measurement $z_t$, the posterior density of $\mathbf{c}^{j_{t}}_{t}$ is 
\begin{align}
 p\left(\mathbf{c}^{j_{t}}_{t}\Big| z_t, \mathbf{v}_{t-1},\mathbf{c}^{j_{t-1}}_{t-1}\right)\propto& p\left(z_t\Big|\mathbf{c}_{t}^{j_t},\mathbf{v}_{t-1},\mathbf{c}^{j_{t-1}}_{t-1}\right)p\left(\mathbf{c}^{j_{t}}_{t}\Big|\mathbf{v}_{t-1},\mathbf{c}^{j_{t-1}}_{t-1}\right)\notag\\
=&p_z\left(z_t\Big|g_m(\mathbf{c}^{j_{t}}_{t})\right)p_v\left(\mathbf{c}^{j_{t}}_{t}\Big|\mathbf{v}_{t-1},\mathbf{c}_{t-1}^{j_{t-1}}\right)
\end{align}

We shall be considering this, in particular,   in the case when the transition is a single time step  and $j_1=j_{t-1}$ and $j_2=j_t$, for a pair $\{j_1,j_2\}\in\bm{\mathcal{I}}\varprod\bm{\mathcal{I}}$, we  use the notation 
\begin{align}
\left\{j_{1}\rightarrow j_{2},p\left(\mathbf{c}^{j_{2}}_{2}\Big| z_2, \mathbf{v}_{1},\mathbf{c}^{j_{1}}_{1}\right)\right\}
\end{align}
for the potential path from $\mathbf{c}^{j_{1}}_{1}$ to $\mathbf{c}^{j_{2}}_{2}$  and its associated conditional probability. 

Similarly, for any pair $\{j_{t-1},j_{t}\}\in\bm{\mathcal{I}}\varprod\bm{\mathcal{I}}$, we can have such a path with its posterior probability as
\begin{align}
\left\{j_{t-1}\rightarrow j_{t},p\left(\mathbf{c}^{j_{t}}_{t}\Big| z_t, \mathbf{v}_{t-1},\mathbf{c}^{j_{t-1}}_{t-1}\right)\right\}
\end{align}

Then, starting from $t=2$, for a fixed $j_{1}$, a local optimal path with greatest posterior probability is given by
\begin{align}
\left\{j_{1}\rightarrow j^*_{2}\Big|\;j^*_{2}=\arg\max_{j_{2}\in\bm{\mathcal{I}}}p\left(\mathbf{c}^{j_{2}}_{2}\Big| z_2, \mathbf{v}_{1},\mathbf{c}^{j_{1}}_{1}\right)\right\}\label{pair1}
\end{align}
Similarly, at $t=3$, for pair $\{j^*_2,j_3\}$, where $j^*_2$ is obtained from~\eqref{pair1} and therefore fixed and $j_3\in\bm{\mathcal{I}}$,  the following local optimal path is
\begin{align}
\left\{j^*_{2}\rightarrow j^*_{3}\Big|\;j^*_{3}=\arg\max_{j_{3}\in\bm{\mathcal{I}}}p\left(\mathbf{c}^{j_{3}}_{3}\Big| z_3, \mathbf{v}_{2},\mathbf{c}^{j^*_{2}}_{2}\right)\right\}\label{pair2}
\end{align}
Iteratively, for any pair $\{j^*_{t-1},j_{t}\}$, $j_{t}\in\bm{\mathcal{I}}$, the following optimization is done:
\begin{align}
\left\{j^*_{t-1}\rightarrow j^*_{t}\Big|\;j^*_{t}=\arg\max_{j_{t}\in\bm{\mathcal{I}}}p\left(\mathbf{c}^{j_{t}}_{t}\Big| z_t, \mathbf{v}_{t-1},\mathbf{c}^{j^*_{t-1}}_{t-1}\right)\right\}\label{pairt}
\end{align}

As a result, for any fixed $j_1\in\bm{\mathcal{I}}$, there is a candidate path that the vehicle potentially  travels:
\begin{align}
&\left\{j_{1}\rightarrow j^*_{2}\rightarrow \cdots\rightarrow j^*_{t}\rightarrow\cdots\rightarrow j^*_{T-1}\rightarrow j^*_{T}\right\}\triangleq\bm{\mathcal{J}}^*_{j_1}\label{seq}
\end{align}
where
\begin{align}
 j^*_{t} = 
  \begin{cases}
    \displaystyle \arg\max_{j_{t}\in\bm{\mathcal{I}}}p\left(\mathbf{c}^{j_{t}}_{t}\Big| z_t, \mathbf{v}_{1},\mathbf{c}^{j_{1}}_{1}\right)&t=2\\
    \displaystyle \arg\max_{j_{t}\in\bm{\mathcal{I}}}p\left(\mathbf{c}^{j_{t}}_{t}\Big| z_t, \mathbf{v}_{t-1},\mathbf{c}^{j^*_{t-1}}_{t-1}\right)&t=3,\cdots,T
  \end{cases}\label{algo}
\end{align}

 Since $\bm{\mathcal{J}}^*_{j_1}$ depends on the selection of $j_1$, we are able to collect the potential  paths corresponding to all possible values of $j_1$, i.e. $\bm{\mathcal{J}}^{*-}=\{\bm{\mathcal{J}}^*_1,\cdots,\bm{\mathcal{J}}^*_{n^2}\}$. Then,
\begin{align}
j^*_{1}&=\arg\max_{\bm{\mathcal{J}}^{*-}}p_z\left(z_1\Big|g_m(\mathbf{c}_1^{j_1})\right)p\left(\mathbf{c}^{j^*_{2}}_{2}\Big| z_2, \mathbf{v}_{1},\mathbf{c}^{j_{1}}_{1}\right)\prod_{t=3}^Tp\left(\mathbf{c}^{j^*_{t}}_{t}\Big| z_t, \mathbf{v}_{t-1},\mathbf{c}^{j^*_{t-1}}_{t-1}\right)\label{algo2}
\end{align}
and the complete estimated path is  \begin{align}
&\left\{j^*_{1}\rightarrow j^*_{2}\rightarrow \cdots\rightarrow j^*_{t}\rightarrow\cdots\rightarrow j^*_{T-1}\rightarrow j^*_{T}\right\}\triangleq\bm{\mathcal{J}}^*\label{seq2}
\end{align}

\begin{figure}[htp!]
    \centering
    \includegraphics[width=1\textwidth]{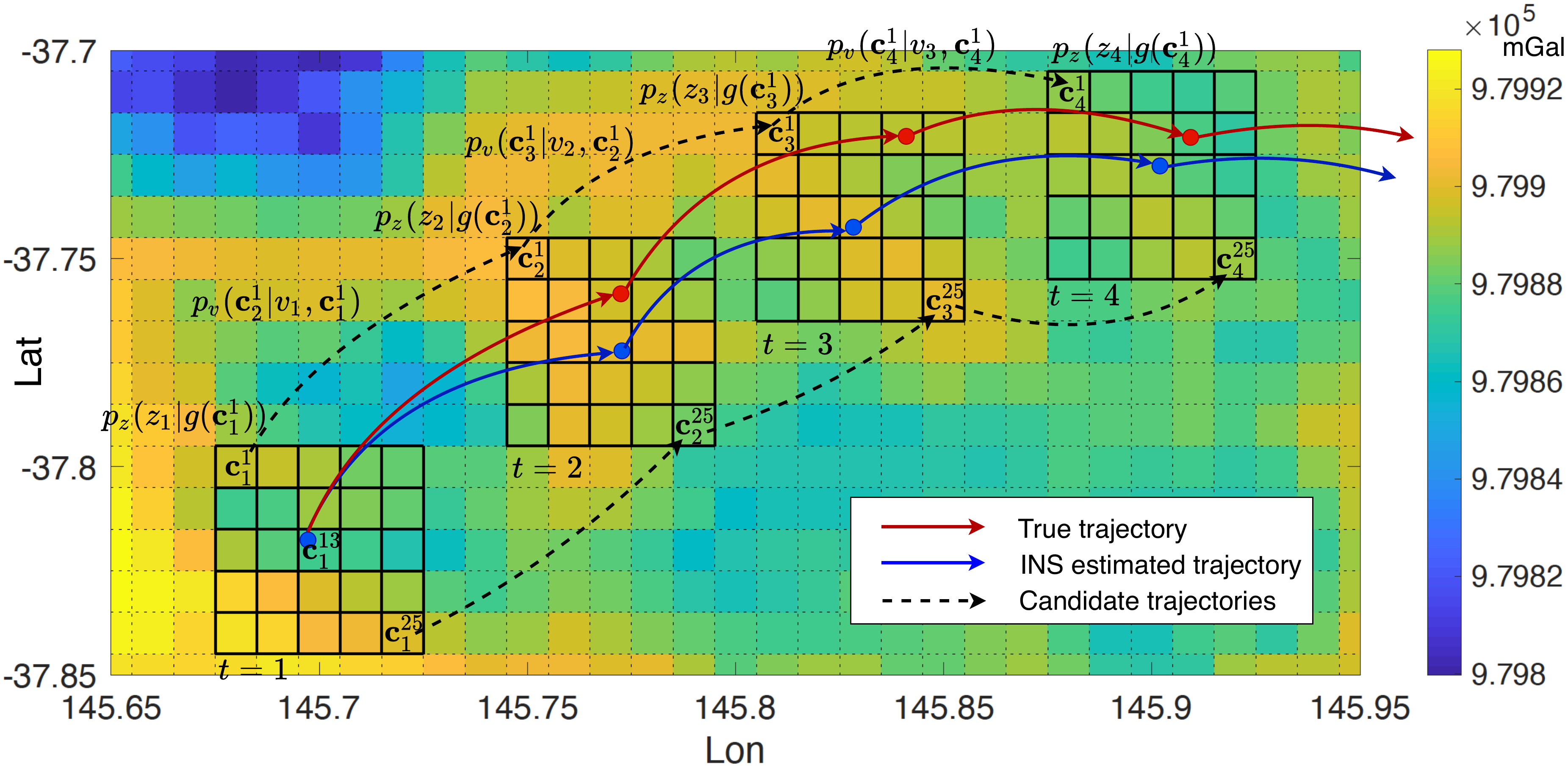}
    \caption{An example of the Viterbi map matching algorithm. The time sequence is $t=1,2,3,4$ and the dimension of each grid  is $5\times5$. The blue lines are the INS estimated positions while the red ones are the actual positions. The black dashed lines indicate two candidate trajectories satisfying~\eqref{seq}, i.e. $\bm{\mathcal{J}}^*_1=\{1\rightarrow 1\rightarrow 1\rightarrow 1\}$ and $\bm{\mathcal{J}}^*_{25}=\{25\rightarrow 25\rightarrow 25\rightarrow 25\}$. All trajectories shown in this figure are for demonstration only.}
    \label{algo_example}
\end{figure}

Eventually, the estimated path $\bm{\mathcal{J}}^*$ is used to reset the INS estimated position directly. 
 An example of how the algorithm works is shown in Fig.~\ref{algo_example}.
 
In practice, given a measurement $z_t$ at a specific time $t$, we may find more than one cell within a grid $\mathbf{B}_t$ that has the same likelihood value using \eqref{eq:likelihood_z}. Therefore multiple optimal trajectories may be identified  because of that they can have the same maximum posterior value. If this happens we choose the trajectory that is closest to the INS estimated position for simplicity.

\subsection{An enhanced two-layer algorithm}\label{subsec_algo_two}
In Section~\ref{subsec_algo}, a grid-based Viterbi algorithm is proposed to find a path used to correct $\mathbf{S}^{INS}_k$. However, it should be noted that the proposed algorithm only returns a path that passes through the center of each cell. As a result, the deviation between this path and the actual trajectory is highly dependent on the resolution of the gravity map since the actual trajectory can be any point of each cell. In this section, an enhanced method is presented to improve the accuracy of the estimated path.

\begin{figure}[htp!]
    \centering
    \includegraphics[scale=.6]{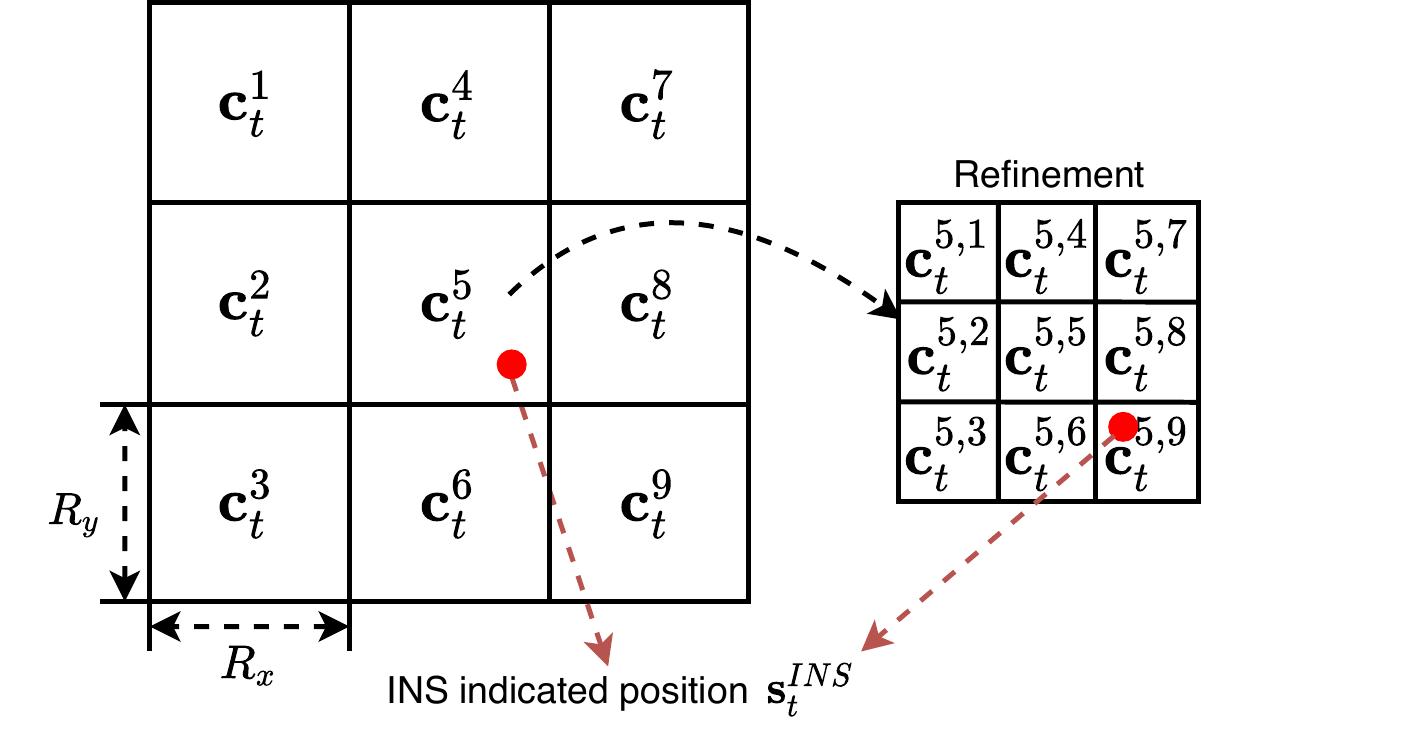}
    \caption{Illustration of the refined grid. The cell $\mathbf{c}_t^5$ is an example to show how the refined sub-cells are constructed, where $n=3$ and $o=3$.}
    \label{grid2}
\end{figure}

The key idea of this enhanced algorithm is to generate refined sub-cells for each cell and use these refined sub-cells to match the actual trajectory. This strategy takes effect when constructing the grid $\mathbf{B}_t$. An example of a refined cell is given in Fig.~\ref{grid2}.

Suppose that each cell is divided into $o\times o$ sub-cells, the centers of which, for $\mathbf{c}_t^{j_t}$, are denoted by $\mathbf{c}_t^{j_t,l_t}$ where $l_t \in \bm{\mathcal{O}}$ and $\bm{\mathcal{O}}=\{1,2,\cdots,o^2\}$. We also use $\mathbf{c}_t^{j_t,l_t}$ as label of $l_t$-th sub-cell of $j_t$-th cell at time $t$. Since the resolution of gravity map is fixed, therefore the sub-cells $\mathbf{c}_t^{j_t,l_t}$ refine cells $\mathbf{c}_t^{j_t}$ to provide better resolution on positions and do not improve the actual resolution of the gravity map. In the other word, the cell $\mathbf{c}_t^{j_t}$ within $\mathbf{B}_t$ is divided into multiple sub-cells $\mathbf{c}_t^{j_t,l_t}$. Then we have the following relation, for any given $j_t$:
\begin{align}
g_m(\mathbf{c}_t^{j_t})=g_m(\mathbf{c}_t^{j_t,l_t}),\;\;\forall l_t\in \bm{\mathcal{O}}
\end{align}
and
\begin{align}
p_z\left(z_t\Big|g_m(\mathbf{c}_t^{j_t,l_t})\right)=p_z\left(z_t\Big|g_m(\mathbf{c}_t^{j_t})\right),\;\;\forall l_t\in \bm{\mathcal{O}}
\end{align}

In similar vein to the~\eqref{pairt},~\eqref{seq} and~\eqref{algo}, the potential path for a given $\{j_1,l_1\}$ is
\begin{align}
\left\{\{j_1,l_{1}\}\rightarrow \{j^*_2,l^*_{2}\}\rightarrow \cdots\rightarrow  \{j^*_{T-1},l^*_{T-1}\}\rightarrow \{j^*_{T},l^*_{T}\}\right\}\triangleq\bm{\mathcal{L}}_{j_1,l_1}^*\label{path_subcell}
\end{align}
where $l^*_{t}$ in $\{j^*_t,l^*_{t}\}$ is the optimal sub-cell and $j^*_t$ is the associated cell.  $\{j^*_t,l^*_{t}\}$ is obtained using $\{j^*_{t-1},l^*_{t-1}\}$ and it can be calculated from~\eqref{algo} via replacing $j_t$ by $\{j_t,l_t\}$. 

Let $\bm{\mathcal{L}}^{*-}=\{\bm{\mathcal{L}}^*_{1,1},\bm{\mathcal{L}}^*_{1,2}\cdots,\bm{\mathcal{L}}^*_{n^2,o^2}\}$. Then $\{j^*_1,l^*_{1}\}$ in the estimated path\begin{align}
    \bm{\mathcal{L}}^*=\left\{\{j^*_1,l^*_{1}\}\rightarrow \{j^*_2,l^*_{2}\}\rightarrow \cdots\rightarrow  \{j^*_{T-1},l^*_{T-1}\}\rightarrow \{j^*_{T},l^*_{T}\}\right\}\label{path_subcell2}
\end{align}
is obtained by
\begin{align}
\{j^*_1,l^*_{1}\}
=\arg\max_{\bm{\mathcal{L}}^{*-}}p_z\left(z_1\Big|g_m(\mathbf{c}_1^{j_1,l_1)}\right)p\left(\mathbf{c}^{j^*_{2},l^*_{2}}_{2}\Big| z_2, \mathbf{v}_{1},\mathbf{c}^{j_{1},l_{1}}_{1}\right)\prod_{t=3}^Tp\left(\mathbf{c}^{j^*_{t},l^*_{t}}_{t}\Big| z_t, \mathbf{v}_{t-1},\mathbf{c}^{j^*_{t-1},l^*_{t-1}}_{t-1}\right)\label{algo3}
\end{align}

\subsection{Reducing the Computation}\label{subsec_algo_reduce}
In the algorithms presented above, the  computational overhead increases dramatically as more cells are considered in the algorithm. To mitigate this issue, we constrain the number of $j_t$ in each time $t$ via a threshold. Relative likelihood values to select the available $j_t$ are calculated as follows: \begin{align}\label{select_likelihood}
\bm{\mathcal{A}}_t =\left\{j_t\left| \frac{p_z\left(z_t\Big|g_m(\mathbf{c}_t^{j_t})\right)}{\max_{j_t\in\bm{\mathcal{I}}} p_z\left(z_t\Big|g_m(\mathbf{c}_t^{j_t})\right)}\geq \alpha,\forall j_t\in\bm{\mathcal{I}}\right.\right\}
\end{align}
where $\alpha\in[0,1]$. \eqref{select_likelihood} implies that only those cells with relatively high likelihood values are chosen. In other words, cells that are very unlikely to form part of the path are discarded. 

\begin{figure}[htp!]
    \centering
    \includegraphics[width=1\textwidth]{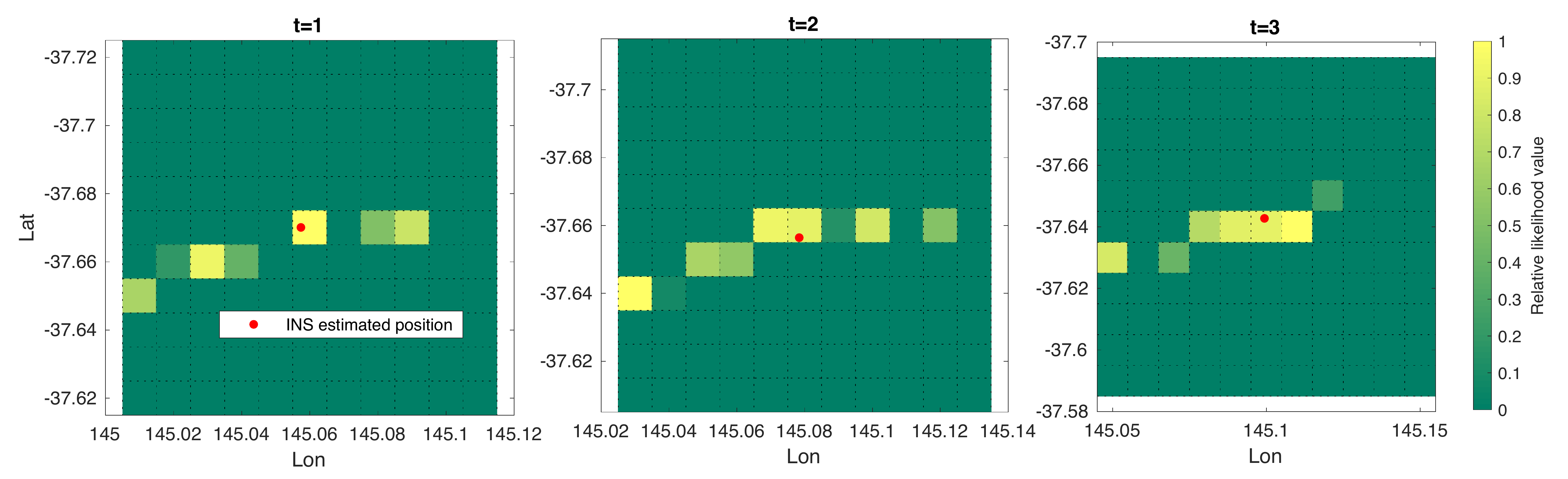}
    \caption{Illustration of likelihood values of cells within  the grids at successive time stamp $t=1,2,3$. The windows size is $n=11\times 11$. For arbitrary cell $j_t$ at time $t$, the likelihood values are calculated from $p_z(z_t|g_m(\mathbf{c}_t^{j_t}))$ and normalized via ${p_z(z_t|g_m(\mathbf{c}_t^{j_t}))}\big/{\max_{j'_t\in\bm{\mathcal{I}}}\left\{p_z(z_t|g_m(\mathbf{c}_t^{j'_t}))\right\}}$. There is no sub-cell for each cell to simplify the illustration. In each grid, there are only up to $7$ cells having relative high likelihood values while other cells' values are close to $0$. If \eqref{select_likelihood} is applied and  let $\alpha=0.1$, the maximum number of possible trajectories is $7^3$; otherwise, in the algorithm without applying \eqref{select_likelihood}, the maximum number of possible trajectories is $121^3$.}
    \label{likelihood_fig}
\end{figure}

An example of higher likelihood  cells from $\mathbf{c}_t^{j_t}$, $j_t\in\bm{\mathcal{I}}$, is given in Fig.~\ref{likelihood_fig}, where we assume that $T=3$ and the size of the grid at each time stamp is $n=11$. The values of $p_z\left(z_t\Big|g_m(\mathbf{c}_1^{j_t})\right)$, $t=1,2,3$, are represented in different colors.  At time $1$, there are three higher likelihood cells. Similarly, at time 2 and 3, there are $3$ and $2$ grids available respectively. As a result, $\bm{\mathcal{A}}_t$, $t=1,2,3$, is obtained accordingly, e.g. $\bm{\mathcal{A}}_1=\{8,18,29,40,61,83,94\}$. The computational complexity is reduced from $O((T-1)n^2)$ to $O(\sum_{t=1}^{T-1} \#\bm{\mathcal{A}}_t\times\#\bm{\mathcal{A}}_{t+1})$, where $\#\bm{\mathcal{A}}_t$ is the number of elements in $\bm{\mathcal{A}}_t$.
In practice,  to simplify the problem, $\alpha$ can be set to be a fixed value, such as $0.1$ or $0.2$.  

Similarly, we  use indexes in ${\bm{\mathcal{A}}_1,\cdots,\bm{\mathcal{A}}_T}$ to find multiple candidate paths using the same method as~\eqref{path_subcell} by replacing the associated indexes sequence. i.e.
\begin{align}
\left\{\{j_1,l_{1}\}\rightarrow \{j^*_2,l^*_{2}\}\rightarrow \cdots\rightarrow  \{j^*_{T-1},l^*_{T-1}\}\rightarrow \{j^*_{T},l^*_{T}\}\right\}\triangleq\bm{\mathcal{R}}_{j_1,l_1}^*\label{seq_likelihood}
\end{align}
where $j_1\in\bm{\mathcal{A}}_1$ and $j^*_t\in\bm{\mathcal{A}}_t$ for $t=2,\cdots,T$.

Let $\bm{\mathcal{R}}^{*-}=\left\{\bm{\mathcal{R}}_{j_1,l_1}^*\big|\;\forall j_1\in\bm{\mathcal{A}}_1\right\}$ contain these candidate paths. Then~\eqref{algo3} can be rewritten as, where $j_1\in\bm{\mathcal{A}}_1$,
\begin{align}
\{j^*_1,l^*_{1}\}
=\arg\max_{\bm{\mathcal{R}}^{*-}}p_z\left(z_1\Big|g_m(\mathbf{c}_1^{j_1,l_1})\right)p\left(\mathbf{c}^{j^*_{2},l^*_{2}}_{2}\Big| z_2, \mathbf{v}_{1},\mathbf{c}^{j_{1},l_{1}}_{1}\right)\prod_{t=3}^Tp\left(\mathbf{c}^{j^*_{t},l^*_{t}}_{t}\Big| z_t, \mathbf{v}_{t-1},\mathbf{c}^{j^*_{t-1},l^*_{t-1}}_{t-1}\right)\label{algo4}
\end{align}

The estimated path is, then, as follows:
\begin{align}
    \bm{\mathcal{R}}^*=\left\{\{j^*_1,l^*_{1}\}\rightarrow \{j^*_2,l^*_{2}\}\rightarrow \cdots\rightarrow  \{j^*_{T-1},l^*_{T-1}\}\rightarrow \{j^*_{T},l^*_{T}\}\right\},\;\;\text{where}\;\;j_t\in\bm{\mathcal{A}}_t\label{path_likelihood}
\end{align}

In summary, the algorithm proposed in Sections~\ref{subsec_algo} with enhanced strategies presented in~\ref{subsec_algo_two} and~\ref{subsec_algo_reduce} is given in Algorithm~\ref{algo_summary}.
\begin{algorithm}
\SetAlgoLined
\KwResult{The estimated trajectory}
 Initialization
  $k\gets 1$
  
 \While{ $k<\text{Running time}$}{
    Obtain position via INS
    
    Push measurements and INS estimated position into vector $\mathbf{Z}_k$, $\mathbf{S}^{INS}_k$ and $\mathbf{V}_k$ respectively. 
    
    \If{$k \mod{T}==0$}
    {Construct grids $\mathbf{B}_t$ using $n$, $o$,  \eqref{eq1} and sub-cell method proposed in Section~\ref{subsec_algo_two} for each $t=1,\cdots,T$
    
    Find all cells $\bm{\mathcal{A}}_t$ satisfying~\eqref{select_likelihood} for each $t=1,\cdots,T$
    
    \For{$j_1\in\bm{\mathcal{A}}_1$ }{
    \For{$l_1\gets1$ \KwTo $o$ }{ 
    \For{$t\gets1$ \KwTo $T$ }{
    Find ordered sequence $\bm{\mathcal{R}}_{j_1}^*$ according to~\eqref{seq_likelihood}
    }
    Push $\bm{\mathcal{R}}_{j_1}^*$ into $\bm{\mathcal{R}}^{*-}$
    }
    }
    Find the estimated path $\bm{\mathcal{R}}^{*}$ using $\bm{\mathcal{R}}^{*-}$ via~\eqref{path_likelihood}
    
    Replace the INS estimated trajectory using $\bm{\mathcal{R}}^{*}$
    
    Clear $\mathbf{Z}_k$, $\mathbf{S}^{INS}_k$ and $\mathbf{V}_k$
    }
    $k\gets k+1$
  
 }
 \caption{The Viterbi-based map matching algorithm }\label{algo_summary}
\end{algorithm}

\section{Simulation}

\begin{figure}[htb!]
\centering
    \subfloat[][]{\includegraphics[width=.46\textwidth]{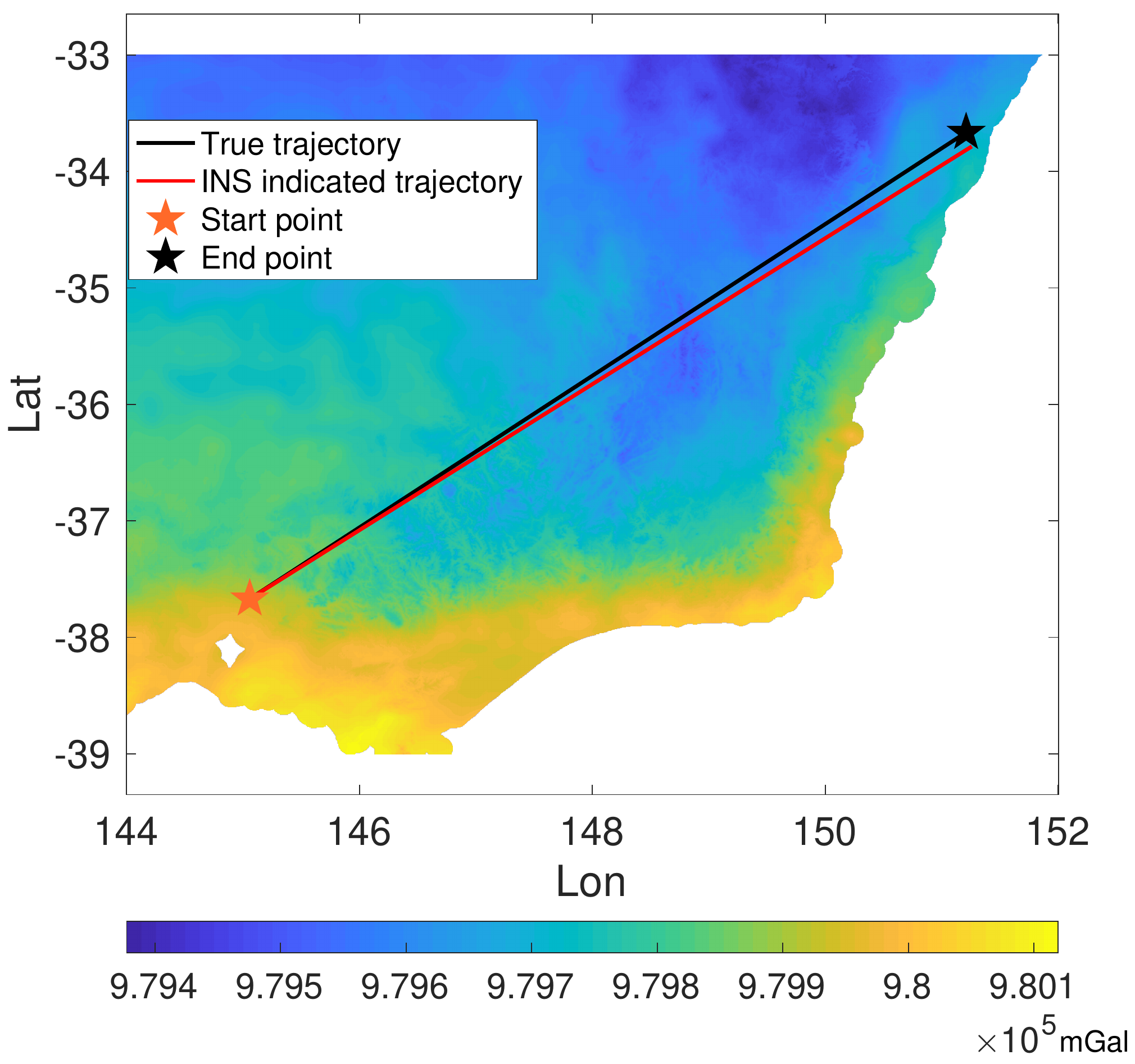}\label{fig:diagram_1}}
\hfil
    \subfloat[][]{\includegraphics[width=.5\textwidth]{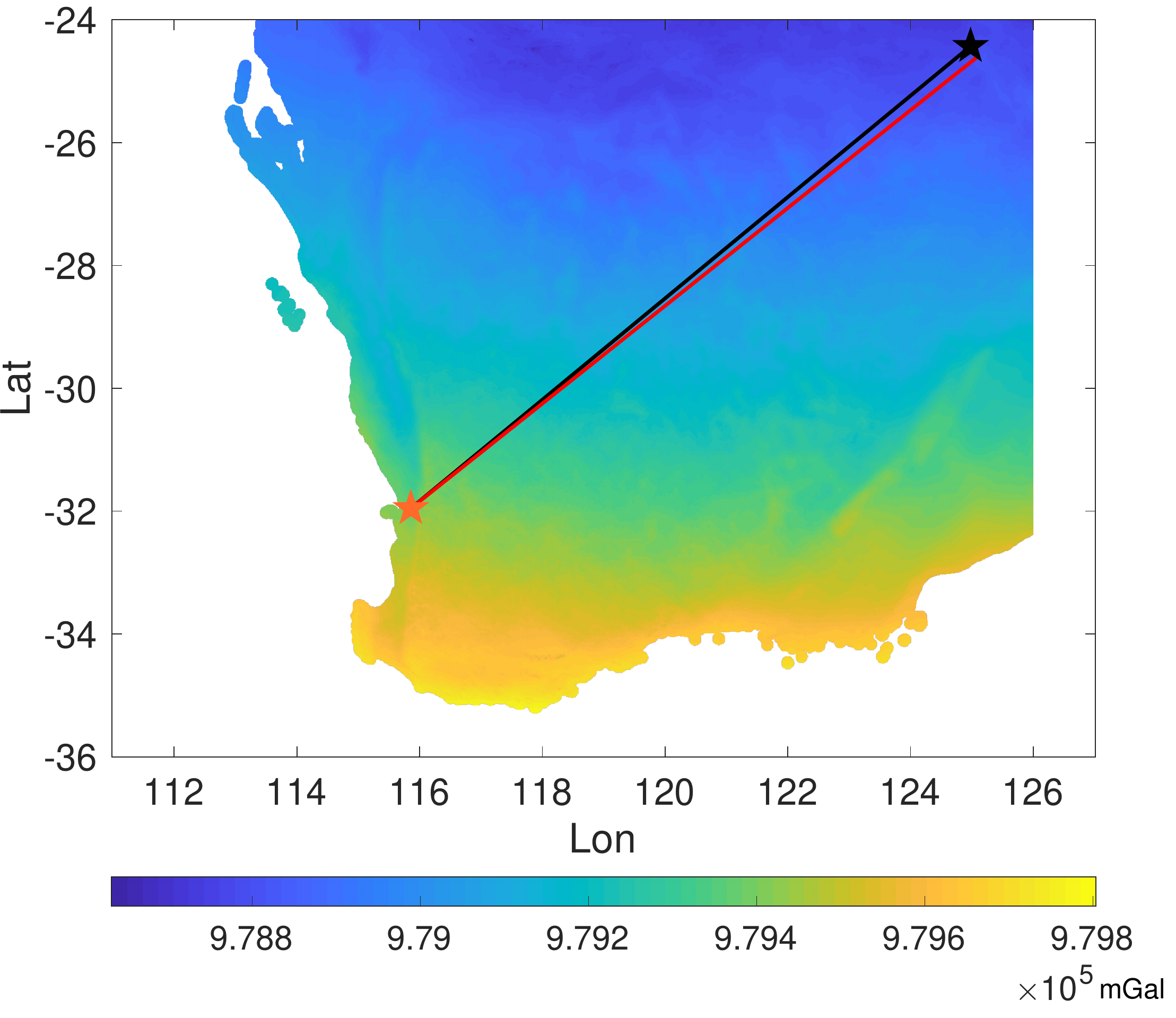}\label{fig:diagram_2}}
    \caption{Illustration of the scenarios used in the simulation.  The colored map is the reference gravitational acceleration map of  Victoria and New south Wales (\ref{fig:diagram_1}); and Western Australia (\ref{fig:diagram_2}).  The INS estimated trajectory without aiding  deviates gradually from the true trajectory. and its resolution is $\frac{1}{200}^\circ$. }
    \label{fig:diagram}
\end{figure}

In the first part of the simulation, we assume that a vehicle (aircraft) travels across Australia,  approximately from Melbourne to Sydney, at a constant velocity $7.54^\circ$/h, roughly  $837$~Km/h ($1^\circ\approx 111\text{Km}$~\citep{scales2017scale}). The total flight distance is about $814$~Km. The gravity map of the  region of interest has been obtained from the GGMplus software~\citep{hirt2013new} with a variety of resolutions, where the component taken from the map is gravitational acceleration. The map is represented in grids that have specified resolutions with the grid equally spaced in terms of geodetic (GRS80) latitude and longitude. The map with lower resolution is created by down-sampling the higher one. An illustrative example is shown in Fig.\ref{fig:diagram_1}

To simplify the simulation, the noise in the velocity measurement, i.e.~\eqref{vel_mea}, from the  IMU is modelled as $\bm{\phi}_{k}=\mathbf{b}_{k}+\mathcal{N}(0,\sigma_v^2\mathbf{I}_2)$, where $\mathbf{b}_{k}$ is bias vector. The measured gravitational acceleration is assumed to be corrupted by additive Gaussian noise $\mathcal{N}(0,\sigma_z^2)$, where $\sigma_z$ is the standard deviation. As mentioned earlier, the true Earth gravity is obtained from gravity map without noise for the purpose of demonstrating the algorithm. 

A Monte Carlo (MC) simulation of 500 runs is  carried out for each scenario and simulation performance at time $k$ is measured by the  root-mean-square error, defined as
\begin{align}\label{error1}
\text{Error}_{k}=\frac{1}{nMC}\sum_{m=1}^{nMT}\text{Dist}\left(\mathbf{s}_{k,m},\mathbf{s}_{k,m}^{C\text{-}INS}\right)
\end{align}
where $nMC$ is the number of MC, $\mathbf{s}_{k,m}$ and $\mathbf{s}_{k,m}^{C\text{-}INS}$ are the position and updated position respectively of the target at time $k$ and $m$-th simulation and $\text{Dist}\left(\cdot,\cdot\right)$ is the Haversine distance (Km) between two points given their longitude and latitude~\citep{gade2010non}. 

The algorithms are evaluated by different parameters. In the simulation, the bias of the velocity is set to be $1^\circ/hr$ and $\sigma_v={9\times10^{-6}}^\circ$/s, roughly $1$~m/s. In the simulation, since we assume that the inertial sensors used in the vehicle INS exhibits white noise and bias, such that the INS computed position deviates from ground truth at the end of journey as shown in Fig. \ref{fig:diagram}. The standard deviation of the gravitational acceleration measurement is $\sigma_z=1,2$~{mGal}. The INS 
reports positions and velocities every $12$~s. 

Firstly, the performances of the  Viterbi based map matching algorithm (VBMP)  proposed in Section~\ref{subsec_algo} and the computation reduced algorithm (RVBMP) proposed in Section~\ref{subsec_algo_reduce} are compared in terms of the time efficiency. In the simulation, standard deviation $\sigma_z=1$~{mGal}, the map resolution is $1/100^\circ$, the correction rate, i.e. $T$, is set to be $6$, window size is $n=13$ and the sub-cell algorithm proposed in~\ref{subsec_algo_two} is not applied. It also can be noticed that the value of $\alpha$ can potentially have impact on the performance of RVBMP; that is, the larger the $\alpha$ is, the more cells will be included in $\bm{\mathcal{A}}_t$ and more time will be required to find a solution. In considering this, the performance of RVBMP using different values of $\alpha$ are compared with VBMP. 
When $\alpha=0$, set $\bm{\mathcal{A}}_t$ will contain all cells and the algorithm becomes VBMP. The simulation results are given in Table \ref{table_alpha}, where TCR stands for time consumption ratio  between the time consumption using a specific $\alpha$ and a referenced time, i.e. $ \frac{\text{Time}_{\alpha}}{\text{Time}_{\text{ref}}}$, 
and $\text{Time}_{\alpha=0.4}$ is chosen as the reference in Table \ref{table_alpha}, Mean and Std. Dev. are sample mean and sample standard deviation of $\text{Error}_{k}$, i.e.

\begin{align}
    \text{Mean}=\frac{1}{L}\sum_{k=1}^L \text{Error}_{k},\;\;\;
    \text{Std. Dev.}=\sqrt{\frac{1}{L-1}\sum_{k=1}^L \left(\text{Error}_{k}-\text{Mean}\right)}\notag
\end{align}
where $L$ is the total number of sampling points over the trajectory.
\begin{table}[htb!]
\centering
\caption{Performance comparison between RVBMP and RVBMP using different values of $\alpha$. The scenario is shown in Fig.~\ref{fig:diagram_1}.}
\begin{tabular}{lccc}
\firsthline
\multicolumn{4}{c}{VBMP and RVBMP} \\
\cline{1-4}
$\alpha$  & Mean (Km)&Std. Dev. (Km)&TCR\\
\hline
$0$ (VBMP)&  $1.2802$   & $0.5011$&$35.3817$   \\
$0.05$      &  $1.2890$   & $0.5101$&$2.6162$   \\
$0.1$      &  $1.2786$  &  $0.5219$&$2.0119$    \\
$0.2$      &  $1.5114$  &  $0.5470$&$1.5049$    \\
$0.3$      &  $1.7097$  &  $0.5655$&$1.1986$    \\
$0.4$      &  $2.3764$  &  $0.8590$&$1$    \\
\hline
\lasthline
\end{tabular}\label{table_alpha}
\end{table}

It can be seen that the RVBMP performs better than VBMP in terms of time efficiency and they have the similar tracking performance when $\alpha$ is less than $0.1$.  

\begin{figure}[htp!]
\centering 
    \subfloat[][]{\includegraphics[width=.5\textwidth]{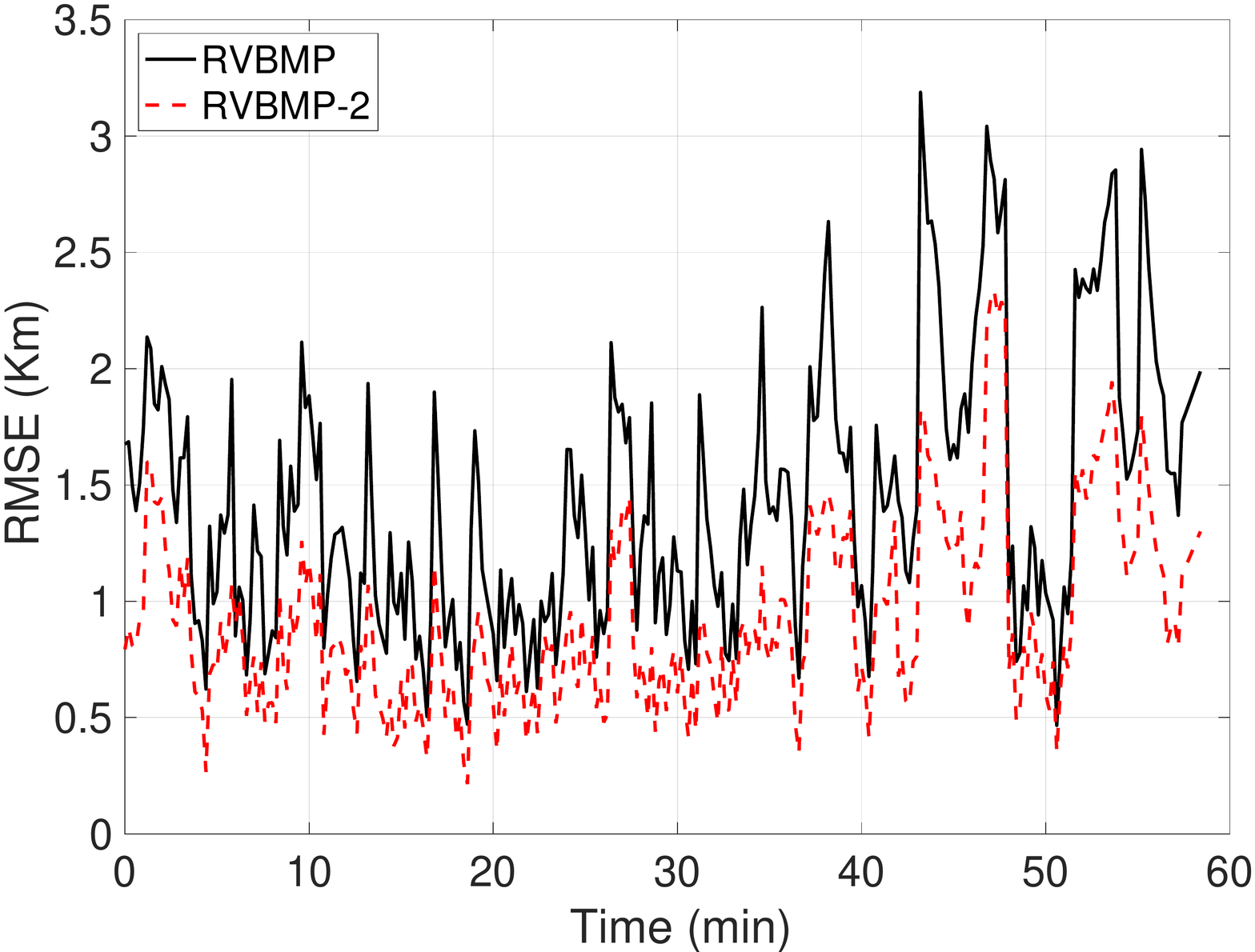}\label{sim00:1}}
\hfil
    \subfloat[][]{\includegraphics[width=.5\textwidth]{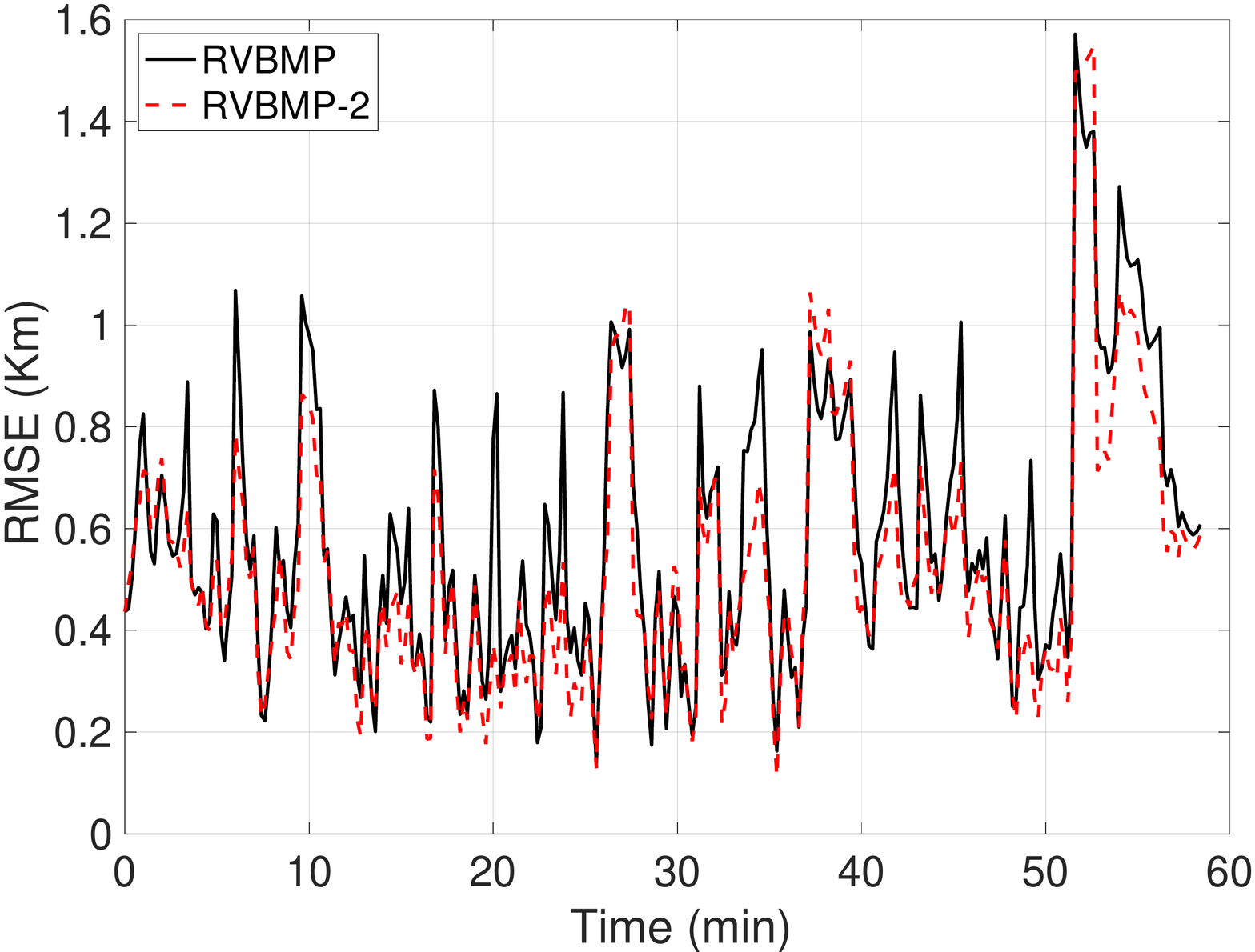}\label{sim00:2}}
    \caption{Simulation results. The scenario is shown in Fig.~\ref{fig:diagram_1}.  \protect\subref{sim00:1} The map resolution is $1/100^\circ$; \protect\subref{sim00:2} The map resolution is $1/200^\circ$.}
    \label{sim00}
\end{figure}

Secondly, the performance of RVBMP  and the  enhanced two-layer algorithm (RVBMP-2) in Section~\ref{subsec_algo_two} are compared via standard deviation $\sigma_z=1$~{mGal} and two map resolutions. The correction rate, i.e. $T$, is set to be $6$, window size is $n=13$ and the size of a sub-cell is $o=7$ for RVBMP-2. The results are given in Fig.~\ref{sim00} and Table~\ref{table_simu0}. The definitions of Mean and Std. Dev are same as those in Table \ref{table_alpha} and they are calculated over those estimated results that are not divergent. The time consumption ratio (TCR) is given to compare the time efficiency of these two algorithms. The TCR in this table is defined as: $\frac{\text{Time}_{RVBMP-2}}{\text{Time}_{RVBMP}}$.

From Fig.~\ref{sim00} and Table~\ref{table_simu0}, we can see that the RVBMP-2 outperforms RVBMP in both resolutions, which indicates that the sub-cell can improve the RVBMP. It should  be noticed that the sub-cell strategy works better when the map resolution is relative low. This is  because of that the high map resolution implies that high position resolution as well so that the sub-cell strategy has less influence on the performance.

\begin{table}[htb!]
\centering
\caption{Performance comparison between RVBMP and RVBMP-2 using two different map resolutions. The scenario is shown in Fig.~\ref{fig:diagram_1}.}
\begin{tabular}{cccc}
\firsthline
\multicolumn{3}{c}{Map resolution = $1/100^\circ$} \\
\cline{1-4}
Algorithm  & Mean (Km)&Std. Dev. (Km)&TCR\\
\hline
RVBMP      & $1.4358 $   & $0.5715$&$1$   \\
RVBMP-2      &  $0.9190$  &  $0.4004$&$24.9$    \\
\hline
\multicolumn{3}{c}{Map resolution = $1/200^\circ$} \\
\cline{1-4}
Algorithm   & Mean (Km)&Std. Dev. (Km)&TCR\\
\hline
RVBMP     &  $0.5846$   & $0.2639$&$1$   \\
RVBMP-2       &  $0.5266$  &  $0.2538$&$28.5$    \\
\hline
\lasthline
\end{tabular}\label{table_simu0}
\end{table}

\begin{figure}[htb!]
\centering
    \subfloat[][]{\includegraphics[width=.5\textwidth]{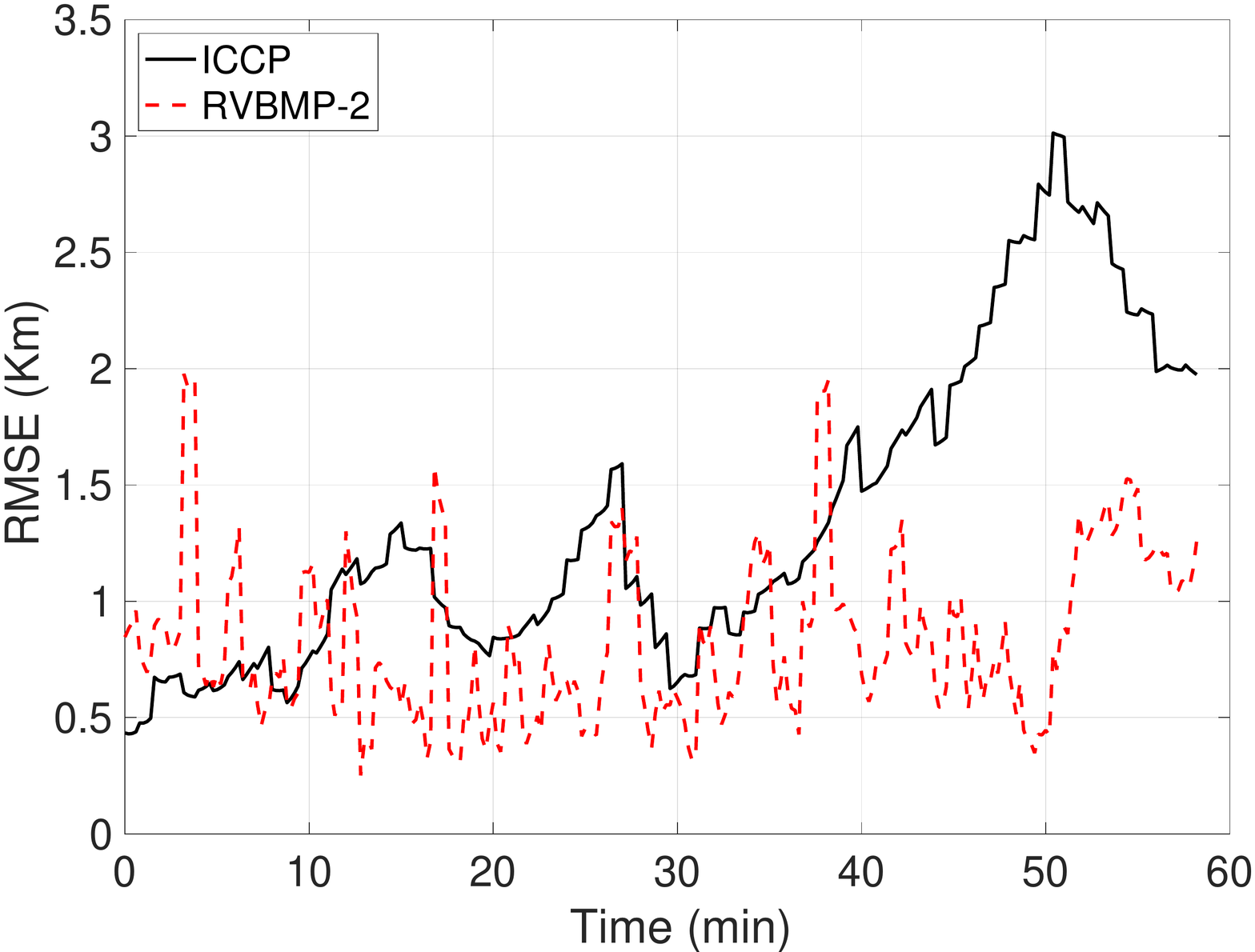}\label{sim1:1}}
\hfil
    \subfloat[][]{\includegraphics[width=.5\textwidth]{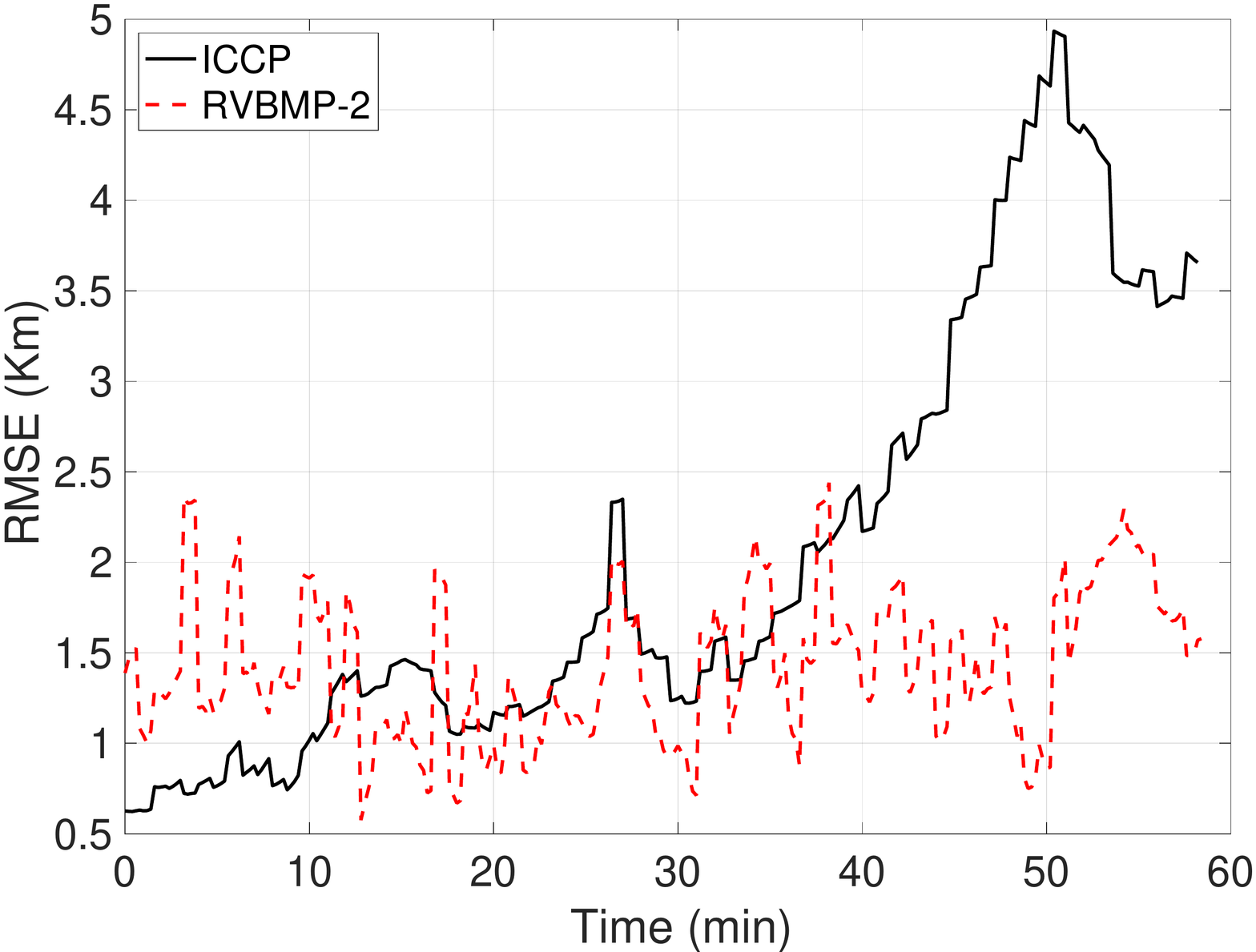}\label{sim1:2}}
    \caption{Simulation results of ICCP and proposed algorithm when  the correction rate is $4$. The scenario is shown in Fig.~\ref{fig:diagram_1}.  \protect\subref{sim1:1} $\sigma_z=1\; \text{mGal}$; \protect\subref{sim1:2} $\sigma_z=2\; \text{mGal}$.}
    \label{sim1}
\end{figure}

\begin{figure}[htb!]
\centering
    \subfloat[][]{\includegraphics[width=.5\textwidth]{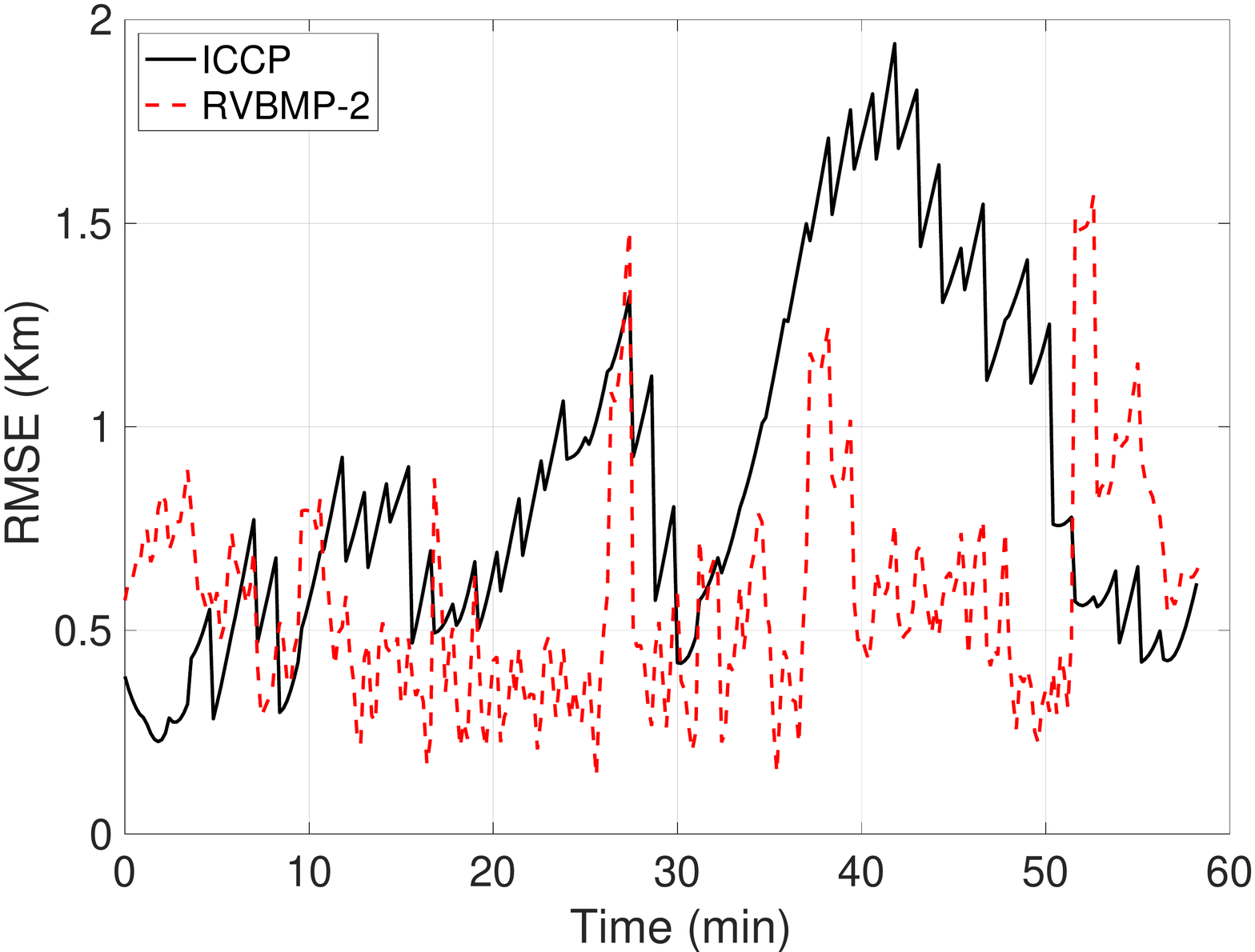}\label{sim2:1}}
\hfil
    \subfloat[][]{\includegraphics[width=.5\textwidth]{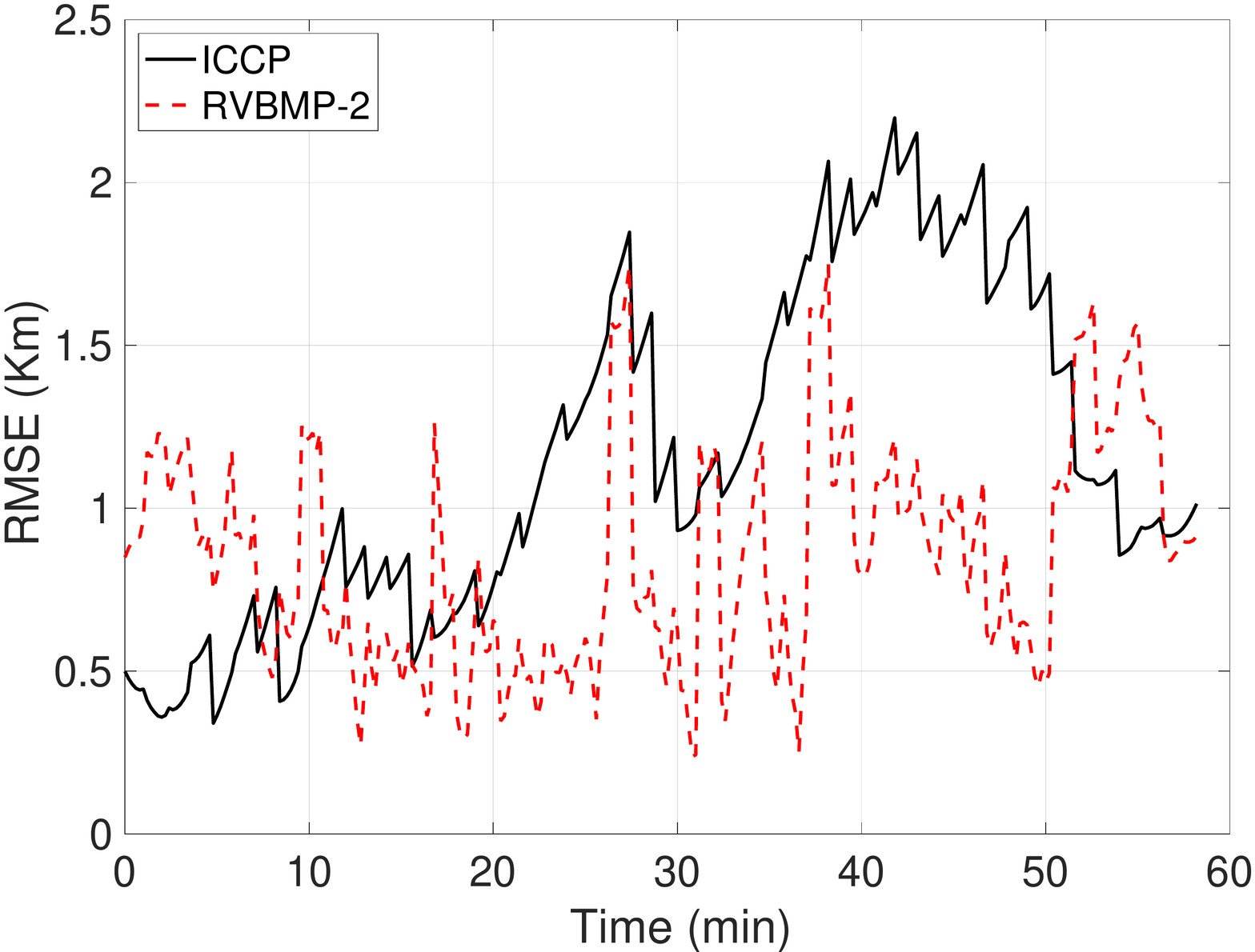}\label{sim2:2}}
    \caption{Simulation results of ICCP and proposed algorithm when the  correction rate is $6$. The scenario is shown in Fig.~\ref{fig:diagram_1}.   \protect\subref{sim2:1} $\sigma_z=1\; \text{mGal}$; \protect\subref{sim2:2} $\sigma_z=2\; \text{mGal}$.}
    \label{sim2}
\end{figure}

Finally, to further demonstrate the efficiency of RVBMP-2, the performance of RVBMP-2 and ICCP~\citep{kamgar1999vehicle,han2017combined} are compared using two different noise levels, i.e. $\sigma_z$. The map resolution is $1/200^\circ$,  window size is $n=13$ and the size of sub-cell is $o=5$. The simulation results are shown in Fig.~\ref{sim1}-\ref{sim2} and summarized in Table~\ref{table_simu}, where the success rate represents the percentage of times the algorithm followed the true trajectory without divergence, and the definitions of Mean and Std. Dev. are as in Table~\ref{table_simu0}.

\begin{table}[htb!]
\centering
\caption{Performance comparison between RVBMP-2 and ICCP using two different standard deviations of the noise, $\sigma_z$ and correction rate. The scenario is shown in Fig.~\ref{fig:diagram_1}.}
\begin{tabular}{cccccc}
\firsthline
\multicolumn{5}{c}{Correction rate = $4$} \\
\cline{1-5}
$\sigma_z$ ({mGal})& Algorithm &Success rate  & Mean (Km)&Std. Dev. (Km)\\
\hline
\multirow{2}{*}{1}&ICCP& $92\%$      & $1.3424$   & $0.6642$       \\
& RVBMP-2&$100\%$       &  $0.8300$     & $0.3557$    \\    
\hline
\multirow{2}{*}{2}&ICCP& $54\%$      &  $2.0253$  &  $1.1920$    \\        
& RVBMP-2&$100\%$       &  $1.4238  $    &$ 0.4146$    \\      
\hline   
\multicolumn{5}{c}{Correction rate = $6$} \\
\cline{1-5}
$\sigma_z$ ({mGal})& Algorithm &Success rate  & Mean (Km)&Std. Dev. (Km)\\
\hline
\multirow{2}{*}{1}&ICCP& $75\%$      & $0.8675$ &  $0.4297$   \\
& RVBMP-2&$100\%$       &  $0.5765$  &  $0.2722$    \\
\hline
\multirow{2}{*}{2}&ICCP& $46\%$      &  $1.1533$   & $0.5148$   \\
& RVBMP-2&$100\%$       &  $0.8478$  &  $0.3466$    \\
\hline
\lasthline
\end{tabular}\label{table_simu}
\end{table}

From Table~\ref{table_simu} and Fig.~\ref{sim1}-\ref{sim2}, we  see that RVBMP-2 algorithm outperforms  ICCP  both in success rate and tracking performance. 

In the second part of the simulation, a reference gravity map that covers a major area of Western Australia (WA), Australia, see Fig.~\ref{fig:diagram_2}, is used to further demonstrate the performance of the proposed algorithm. The major challenge of  this scenario is that the gravitational acceleration is relatively flat; that is, the variation in  gravity across distinct pixels/cells is, on the whole,  smaller than that in the  regions in the earlier simulations. This flatness of the gravity map suggests that aiding with it will   perform less well than in the previous results. 

In the simulation, the start point of the vehicle is near Perth,  and the end point is located at  ($[-24.427111,124.978610]$). The distance is about $1206$~Km. The remaining parameters are similar to those in Table~\ref{table_simu}.

\begin{table}[htb!]
\centering
\caption{Performance comparison between RVBMP-2 and ICCP using two different standard deviations of the noise, $\sigma_z$ and the correction rate. The scenario is shown in Fig.~\ref{fig:diagram_2}.}
\begin{tabular}{cccccc}
\firsthline
\multicolumn{5}{c}{Correction rate = $4$} \\
\cline{1-5}
$\sigma_z$ ({mGal})& Algorithm &Success rate  & Mean (Km)&Std. Dev. (Km)\\
\hline
\multirow{2}{*}{1}&ICCP& $23\%$      & $1.5880$   & $0.8657$       \\
& RVBMP-2&$100\%$       &  $1.4861$     & $0.6183$    \\    
\hline
\multirow{2}{*}{2}&ICCP& $2\%$      &  $2.4581$  &  $1.0113$    \\        
& RVBMP-2&$100\%$       &  $2.3573$    &$0.8325$    \\      
\hline   
\multicolumn{5}{c}{Correction rate = $6$} \\
\cline{1-5}
$\sigma_z$ ({mGal})& Algorithm &Success rate  & Mean (Km)&Std. Dev. (Km)\\
\hline
\multirow{2}{*}{1}&ICCP& $22\%$      & $1.0967$ &  $0.4457$   \\
& RVBMP-2&$100\%$       &  $1.0679$  &  $0.4336$    \\
\hline
\multirow{2}{*}{2}&ICCP& $2\%$      &  $1.6697$   & $0.7551$   \\
& RVBMP-2&$99\%$       &  $1.6003$  &  $0.5975$    \\
\hline
\lasthline
\end{tabular}\label{table_simu2}
\end{table}

In the Table~\ref{table_simu2}, we can see that, in terms of accuracy, the performances of RVBMP-2 and ICCP are similar because of the map is relatively flat and provides less information to the algorithm. However,  the former has better a success rate than the latter. 

\section{Summary}
In this paper, we have presented a Viterbi-inspired map matching algorithm based on a hidden Markov model (HMM) formalisation of the map matching problem. Our algorithm determines the optimal trajectory of a platform given a sequence of gravitational measurements, a geo-reference gravity map and INS velocity information. This trajectory is optimal in the maximum a posterior sense and is used to correct the platform's INS. The simulation results demonstrate that the  algorithm is efficient in correcting the INS estimated  trajectory.  In addition, we proposed an enhanced 2-layer addition to the algorithm to improve the performance when the map resolution is low. We showed by simulations that our proposed algorithm outperforms existing map matching algorithms in terms of navigational accuracy and is also robust to varying levels of sensor noise. In terms of future work, we have validated our algorithm using gravitational measurements however it can be potentially generalized to other aiding resources, such as terrain elevation and magnetic field, that have similar maps. Additionally, we currently use a straightforward method when combining the output of the algorithm with the INS position -- the INS position is simply reset with the corrected position. Accordingly, another area of future work is to develop a more sophisticated  fusion process. 

\section{Conflict of interest} None
\bibliographystyle{agsm}
\bibliography{references}

\end{document}